\documentclass[runningheads]{llncs}
\usepackage[T1]{fontenc}
\usepackage{graphicx}
\begin{document}
\title{Adapting SAM 2 for Visual Object Tracking:\\ 1st Place Solution for MMVPR Challenge Multi-Modal Tracking}

\author{
Cheng-Yen Yang\inst{1}$^{*}$    \and
Hsiang-Wei Huang\inst{1}$^{*}$ \and
Pyong-Kun Kim\inst{2}$^{*}$      \and
Chien-Kai Kuo\inst{1}    \and
Jui-Wei Chang\inst{1}     \and
Kwang-Ju Kim\inst{2}     \and
Chung-I Huang\inst{3}      \and
Jenq-Neng Hwang\inst{1}
}

\authorrunning{C. Yang et al.}
%
\institute{University of Washington, Seattle WA, USA \and
Electronics and Telecommunications Research Institute, Daejeon, South Korea\\ \and
National Center for High-performance Computing, Hsinchu, Taiwan}
\maketitle              
\begin{abstract}
We present an effective approach for adapting the Segment Anything Model 2 (SAM2) to the Visual Object Tracking (VOT) task. Our method leverages the powerful pre-trained capabilities of SAM2 and incorporates several key techniques to enhance its performance in VOT applications. By combining SAM2 with our proposed optimizations, we achieved a first place AUC score of 89.4 on the 2024 ICPR Multi-modal Object Tracking challenge, demonstrating the effectiveness of our approach. This paper details our methodology, the specific enhancements made to SAM2, and a comprehensive analysis of our results in the context of VOT solutions along with the multi-modality aspect of the dataset.

\end{abstract}

\section{Introduction}

\newcommand\blfootnote[1]{%
  \begingroup
  \renewcommand\thefootnote{}\footnote{#1}%
  \addtocounter{footnote}{-1}%
  \endgroup
}

\blfootnote{* These authors are the core members of the challenge team: uwipl and consider equal contribution to the work.}

Visual Object Tracking (VOT) is a core problem in computer vision, where the objective is to detect and continuously track the position of a target object in a sequence of video frames. Despite notable advancements in the field, VOT still faces significant challenges due to various factors such as occlusion, where the object becomes temporarily hidden from view; motion blur, caused by rapid movement of the object or camera; and target deformation, where the object changes shape, size, or appearance over time. These factors make maintaining robust and accurate tracking throughout the video sequence highly difficult.

Traditional tracking approaches, such as discriminative correlation filters (DCF) and Siamese networks, have contributed to significant progress in the field. However, these methods often fail in handling complex scenarios like long-term occlusion or severe target appearance changes. To address these issues, hybrid methods that combine tracking and segmentation, such as SiamMask \cite{siammask}, have been introduced. These approaches refine the object’s boundaries frame by frame but still suffer from inconsistencies when objects reappear after occlusion or undergo dramatic appearance shifts.

Recent developments in segmentation models, particularly the Segment Anything Model 2 (SAM2) \cite{sam2}, have opened new possibilities for enhancing the robustness and efficiency of VOT systems. SAM2, developed by Meta AI, builds upon its predecessor by incorporating real-time video segmentation capabilities, memory modules for tracking, and the ability to handle occlusions and appearance variations with minimal user input. This makes SAM2 a strong candidate for addressing some of the most pressing challenges in VOT.

In this report, we explore the adaptation of SAM2 for the VOT task, leveraging its superior segmentation capabilities to handle complex tracking scenarios. SAM2’s ability to consistently track objects even in occluded or fast-moving scenes makes it highly suitable for VOT. However, the basic segmentation functionality of SAM2 is not sufficient for optimal performance in real-world tracking applications, which often involve variable conditions and dynamic environments. Therefore, we introduce a series of novel techniques specifically designed to enhance SAM2’s performance in VOT.

The proposed tricks include backward tracking and tracklet interpolation. Specifically, our approach achieved an AUC score of 89.4 on the 2024 ICPR Multi-modal Object Tracking Challenge, earning first place among the participants. This demonstrates the potential of advanced segmentation models like SAM2, combined with task-specific adjustments, to push the boundaries of VOT performance.

\section{Related Work}
\subsection{Visual Object Tracking}
Visual Object Tracking (VOT) is a long-standing challenge in computer vision, where the goal is to detect and follow a target object in consecutive video frames. Traditional tracking algorithms include discriminative correlation filters (DCF) and tracking-by-detection approaches. DCF methods, such as KCF (Kernelized Correlation Filters) and ECO (Efficient Convolution Operators), excel in tracking by applying correlation between frames. However, these approaches are often vulnerable to occlusion, scale variation, and deformation. In contrast, learning-based models, particularly those utilizing convolutional neural networks (CNNs), have significantly advanced VOT performance. Models like SiamFC \cite{siamfc} and SiamRPN \cite{siamrpn} introduced the use of Siamese networks, where a template image of the target is matched with candidate regions in the search frame. While these methods have improved robustness, they still struggle when significant appearance changes occur or when objects leave and re-enter the scene. And other post-processing methods leveraging appearance~\cite{huang2023enhancing,yang2024online,sun2024gta}, motion~\cite{huang2024exploring}, or meta-data~\cite{yang2024sea} are required to enhance the tracking performance.

To tackle these challenges, hybrid method such as SAMURAI~\cite{yang2024samurai}, SiamMask \cite{siammask} combined tracking with segmentation. These approaches refined the tracked object’s boundaries frame-by-frame but still faced issues in handling complex backgrounds and long-term tracking.

Recent advancements in transformer-based architectures have further improved tracking. Methods like TransT\cite{transt} and STARK\cite{stark} integrate attention mechanisms to capture long-range dependencies in video sequences, enhancing robustness against occlusion and appearance variation. However, these methods are computationally expensive and may struggle in real-time applications.

\subsection{Multi-modal Tracking}
Multi-modal tracking involves leveraging data from multiple sensor modalities, such as RGB, thermal, or depth cameras, to enhance object tracking in challenging environments. Traditional object tracking systems often rely solely on RGB data, which can struggle under poor lighting conditions, heavy occlusion, or when the object’s appearance changes significantly. Multi-modal systems aim to overcome these limitations by fusing complementary information from different sensors, thereby improving robustness and accuracy.

In recent years, various approaches have emerged for multi-modal visual object tracking. Many of these methods utilize deep learning architectures that incorporate cross-modal fusion, where the features extracted from different modalities are combined. For example, CRAFT~\cite{kuan} proposed the fusion of radar and camera to boost the tracking performance. DiMP~\cite{zhang2019multi} introduced an end-to-end framework that achieves strong performance in RGB-T tracking, while ViPT~\cite{Zhu_2023_CVPR} proposed a multi-modal object tracking framework that achieves decent performance across multiple multi-modal tracking tasks.

Multi-modal tracking has become increasingly important in real-world applications, such as autonomous driving, surveillance, and medical imaging. For instance, in autonomous vehicles, integrating data from LiDAR, RGB cameras, and radar helps maintain accurate tracking even in adverse weather conditions, where certain sensors may become unreliable.

In the context of the 2024 ICPR Multi-modal Object Tracking challenge, the dataset features RGB, depth and infrared imagery, pushing participants to develop techniques that can intelligently fuse these data streams. Models that effectively combine information from both modalities have shown superior performance, especially in scenarios involving low-light environments or occluded objects.

\subsection{SAM2}
The Segment Anything Model 2 \cite{sam2} is an evolution of the original SAM model, designed for both image and video segmentation. SAM2 builds on the concept of Promptable Visual Segmentation, allowing the user to interactively guide the segmentation process with simple clicks or bounding boxes. This makes SAM2 highly versatile, capable of segmenting a wide variety of objects across different scenarios.

Compared to its predecessor, SAM2 introduces several key innovations that improve its performance in video segmentation, particularly in tracking applications. It integrates a memory module that enables it to track objects consistently across frames, even when objects are temporarily occluded or undergo significant appearance changes.

Additionally, SAM2’s ability to work on diverse video datasets, such as DAVIS \cite{davis} and YouTubeVOS \cite{youtube}, makes it well-suited for the VOT task. This capability is enhanced by training on the SA-V dataset \cite{sam}, which contains over 50,000 videos and millions of frames, providing a rich set of object appearances and transformations.

By integrating SAM2 into the VOT pipeline and introducing task-specific modifications, this work demonstrates the potential for SAM2 to push the boundaries of performance in multi-modal tracking environments.

\begin{figure}[t]
    \centering
    \includegraphics[width=1.0\linewidth]{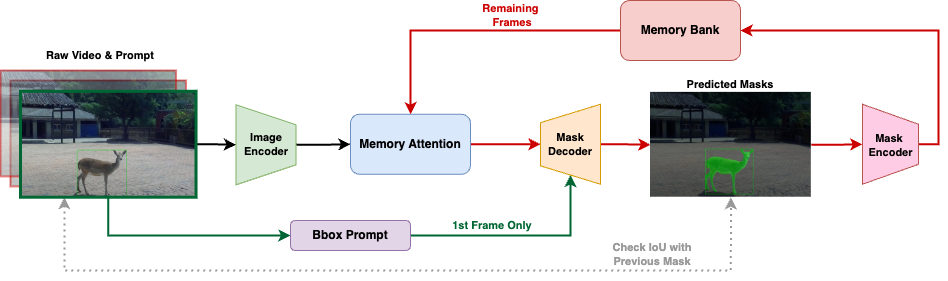}
    \caption{An example illustration of adapting SAM2 \cite{sam2} on VOT task. This pipeline leverages SAM2's powerful segmentation capabilities by using an initial bounding box prompt on the first frame, then utilizes its memory bank feature to propagate and refine object masks through subsequent video frames, enabling efficient and accurate object tracking.}
    \label{fig:enter-label}
\end{figure}

\section{Method}
\subsection{SAM2 for Visual Object Tracking}
In this work, we adapt the Segment Anything Model 2 (SAM2), originally designed for image and video segmentation tasks, to perform Visual Object Tracking (VOT). SAM2’s ability to generate high-quality object masks provides a strong foundation for accurate and robust object tracking. By leveraging its advanced segmentation capabilities, we can consistently detect and localize target objects across a sequence of video frames, even in complex scenarios such as occlusion or rapid movement.

To adapt SAM2 for VOT, we utilize its video object segmentation (VOS) capabilities to segment the target object in each frame. Specifically, SAM2 outputs a segmentation mask that delineates the object of interest. However, VOT traditionally operates with bounding boxes, not segmentation masks. To address this, we extract the bounding box from the segmentation mask by calculating the minimum and maximum coordinates of the mask in each frame.

This approach efficiently translates the pixel-level precision of SAM2’s segmentation masks into the bounding box format typically used in VOT benchmarks, allowing us to achieve high tracking accuracy while maintaining computational efficiency. Through this method, SAM2’s strong segmentation performance is directly leveraged for visual object tracking, making it suitable for a variety of real-world applications.

In addition, the bounding box extraction technique ensures that we can apply SAM2 seamlessly to existing VOT pipelines, without requiring significant modifications to handle segmentation outputs. This combination of segmentation precision and tracking flexibility underpins our approach's success in achieving state-of-the-art performance on multi-modal visual object tracking benchmarks.

\subsection{Backward Tracking}
To enhance the robustness of our tracking system, we introduce Backward Tracking as an additional step in our pipeline to boost the tracking performance for those sequence with poor tracking results. This method involves running the tracking process in reverse after completing the initial forward pass through the video sequence. The rationale behind this approach is that running SAM2 with the final bounding box from the last frame and processing the video in reverse allows the model to leverage different memory states and object features, which may lead to better tracking results in certain challenging frames. The process works as follows:
\begin{enumerate}
    \item Forward Tracking: Initially, we run SAM2 in the normal forward direction on the video, producing a sequence of bounding boxes based on the object’s location in each frame.
    \item Backward Initialization: After obtaining the forward tracking results, we take the last bounding box from the forward pass as the initial object state for the backward pass. This bounding box is used to reinitialize the tracker in reverse.
    \item Backward Tracking Process: We re-run SAM2, but this time we traverse the video frames backward, from the last frame to the first. Since SAM2 operates with a memory module that influences its understanding of the object's position and appearance over time, running the tracker in reverse allows it to have a chance to retrieve missed detections, recover from errors, and achieve improved tracking in certain sequences where the forward pass struggled (e.g., due to occlusion, motion blur, or changes in object appearance).
\end{enumerate}

\subsection{Tracklet Interpolation}

In our adaptation of SAM2 for visual object tracking, we treat each frame as a new instance of object detection. SAM2 identifies the best mask by selecting the prediction with the maximum Intersection over Union (IoU) out of $N_{max}$ predictions for the subsequent frame $t+1$ with the current frame $t$. However, this approach can lead to certain challenges in tracking consistency. Most objects in the dataset exhibit limited deformation and maintain relatively stable bounding box ratios and sizes, given the frame rate and annotation characteristics of the data. Despite this, the frame-by-frame predictions from SAM2 can sometimes result in jittery or inconsistent segmentations, particularly in cases of occlusion or camera movement.

Interestingly, SAM2's memory attention mechanism allows it to successfully track objects with similar appearances across extended sequences of frames. Leveraging this capability, we have developed a post-processing interpolation method to refine these findings and improve tracking stability. 

Our tracklet interpolation method operates on the initial results from SAM2. Let's denote the first pass of SAM2 results as $B = \{b_1, b_2, ..., b_n\}$, where $b_i = (x_i, y_i, w_i, h_i)$ represents the bounding box for frame $i$, with $x_i, y_i$ as the top-left coordinates and $w_i, h_i$ as the width and height respectively.

We calculate the percentage change in the bounding box ratio between consecutive frames:

\begin{equation}
    \Delta r_i = \left|\frac{(w_i/h_i) - (w_{i-1}/h_{i-1})}{w_{i-1}/h_{i-1}}\right| \times 100\%
\end{equation}

\noindent Next, we compute the mean ratio as a thresholding value $t$:

\begin{equation}
    t = \frac{1}{n-1} \sum_{i=2}^n \Delta r_i.
\end{equation}

\noindent We identify sections where the bounding box ratios exhibit extreme changes and label them as "frames to interpolate". These are frames where $\Delta r_i > \alpha t$, where $\alpha$ is a tunable parameter. For each section of frames to interpolate, we use the nearest unaffected frames (frames where $\Delta r_i \leq \alpha t$) as anchor points. Let's denote these anchor frames as $b_a$ and $b_b$. We then interpolate the bounding boxes for frames $i$ where $a < i < b$ using:

\begin{equation}
    b_i' = b_a + \frac{i-a}{b-a}(b_b - b_a)
\end{equation}

This process can be repeated multiple times until convergence or until a satisfactory level of smoothness is achieved. The convergence criterion can be defined as:

\begin{equation}
    \max_{i=2}^n \Delta r_i < \beta t
\end{equation}

\noindent where $\beta$ is another tunable parameter (typically set slightly higher than $\alpha$). This approach allows us to progressively smooth out extreme variations in the bounding box ratios while preserving the overall trajectory of the tracked object.

\begin{figure}[t]
    \centering
    \includegraphics[width=1.0\linewidth]{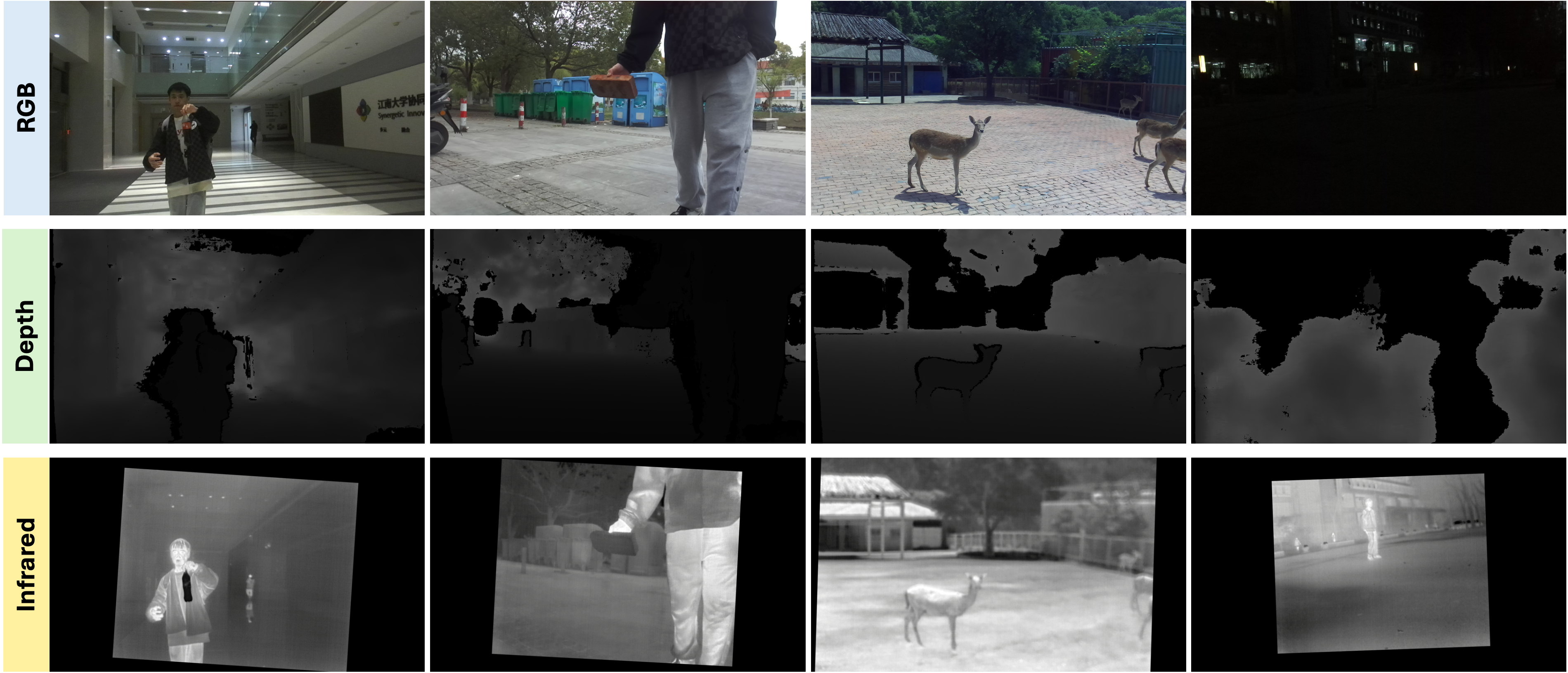}
    \caption{Sample of the multi-modal videos from the testing sequence.}
    \label{fig:dataset}
\end{figure}

\section{Experiments}
\subsection{Dataset}
For our experiments, we utilized the track1 visual object tracking dataset provided by the 2024 ICPR Multi-Modal Visual Pattern Recognition Workshop. This dataset is specifically designed to tackle the challenges of tracking objects across different modalities, including RGB, infrared thermal, depth, and event data. It consists of 500 multi-modal videos, which are divided into 400 videos for training and 100 videos for testing.

It's important to note that although the challenge description mentions 400 videos in the training set, the actual data provided consists of 10 images each along with the annotations for the three modalities (RGB, infrared thermal, and depth). These 10 images appear to be selected randomly from the original videos, which are not provided to the participants.
For the testing set, the length of the sequences varies from 100 to 600 frames. Each test sequence is provided with the ground-truth bounding box for the very first frame, which serves as the initial target location for tracking algorithms.

\begin{figure}[t]
    \centering
    \includegraphics[width=\linewidth]{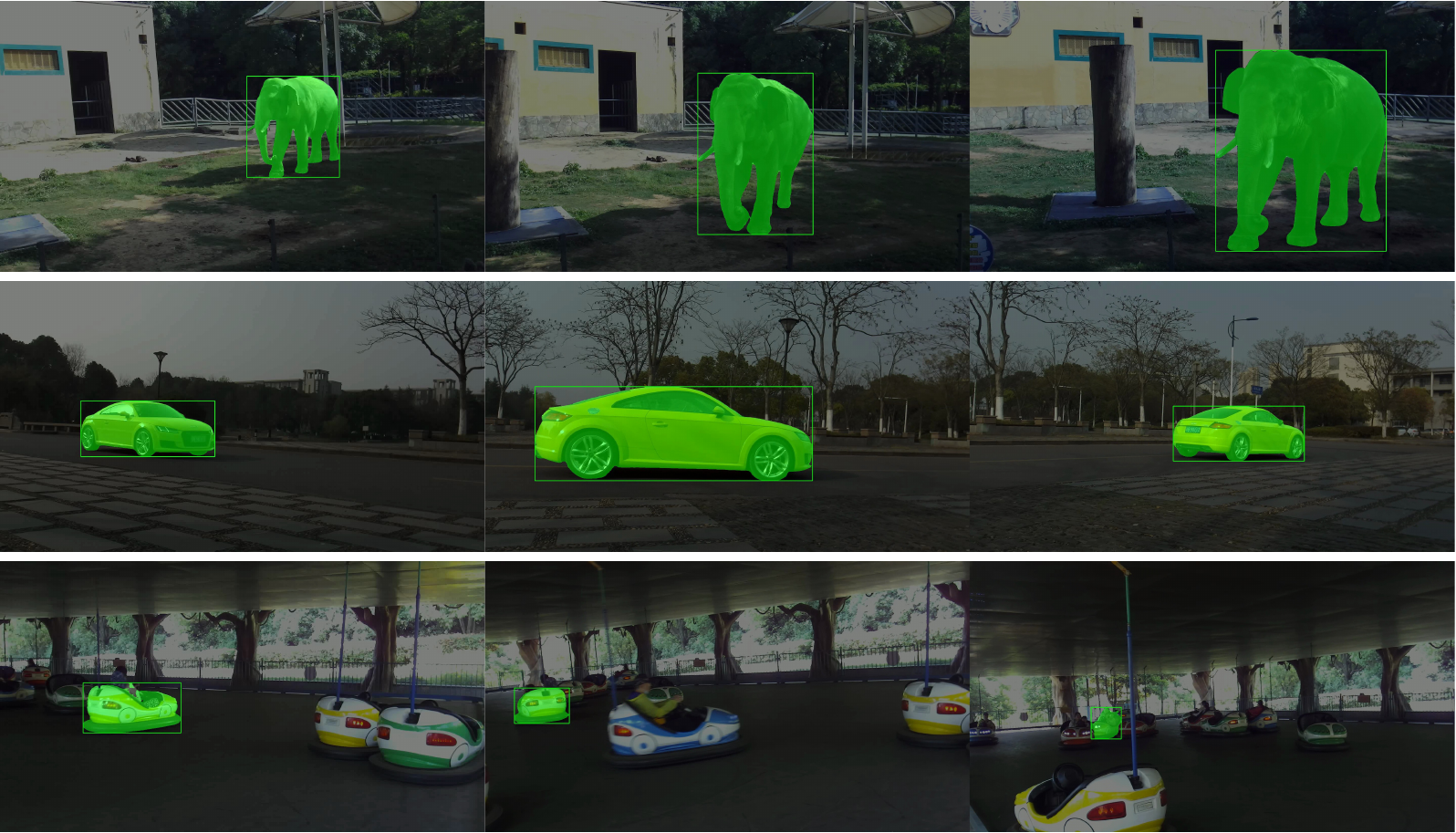}
    \caption{Visualization of our tracking results on the ICPR multi-modal tracking dataset. We selected three different tracking cases with different level of difficulties, caused by the moving speed of object, occlusion, and distractor in the environment.}
    \label{fig:enter-vis}
\end{figure}

\subsection{Metrics}
We utilized the Area Under the Curve (AUC) as our primary evaluation metric. The AUC is calculated based on the success rates at various thresholds of bounding box IoU between groundtruth box and predict box. This metric provides a comprehensive measure of tracking performance across different scenarios.

We calculate the overlap rate between the predicted bounding box ($P_t$) and the ground truth bounding box ($G_t$) in frame $t$:

\begin{equation}
    R_t = \frac{|P_t \cap G_t|}{|P_t \cup G_t|}
\end{equation}

\noindent When the target is visible, $R_t$ measures the IoU value. If the target is out of view or occluded ($G_t$ is empty), $R_t$ is set to 0. We then compute success rates at different thresholds $\theta_i$:

\begin{equation}
    SR(\theta_i) = \frac{1}{N} \sum_{t=1}^N u_t(\theta_i)
\end{equation}

\noindent Where $u_t(\theta_i)$ is 1 if $R_t > \theta_i$, and 0 otherwise. Finally, we calculate the AUC by averaging the success rates across all thresholds:

\begin{equation}
    AUC = \frac{1}{n} \sum_{i=1}^n SR(\theta_i)
\end{equation}

\begin{table}[t]
\centering
\small
\begin{tabular}{ccc}
\hline
Ranking & Team Name          & AUC score (\%) \\ \hline
1     & UWIPL\_ETRI (ours)   & 89.4 \\
2     & xxxxl          & 86.9 \\
3     & Weidlnu        & 86.8 \\
4     & Peace          & 86.8 \\
5     & ylh            & 85.9 \\
6     & hubulai        & 83.1 \\
7     & xxxxxjjjjjxxxx & 81.1 \\
8     & yyy           & 78.5 \\
9     & xuanwang       & 77.8 \\
10    & huxiantao      & 75.9 \\\hline
Baseline & ViPT \cite{vipt} & 74.1 \\\hline
\end{tabular}
\vspace{1em}
\caption{Leaderboard of Track 1 in the Multi-Modal Visual Pattern Recognition 2024: Multi-Modal Tracking. Our method obtained an AUC score of 89.4, ranking in the first-place.}
\label{table:ranking}
\end{table}
\begin{table}[h]
\centering
\small
\begin{tabular}{c|c|c}
\hline
\hspace{1em}Modality\hspace{1em} & \hspace{1em}Model\hspace{1em} & \hspace{1em}AUC score (\%)\hspace{1em} \\ \hline
Depth & SAM2-l & 19.8\\
Infrared & SAM2-l & 56.5\\
RGB & SAM2-l & 88.6\\
\hline

\end{tabular}
\vspace{1em}
\caption{The performance of using SAM2 with different input modality on the 2024 ICPR multi-modal tracking dataset.}
\label{table:multi-modal}
\end{table}

This comprehensive evaluation approach allows us to assess the overall performance of our tracking method across various scenarios and difficulty levels. For the challenge, the AUC is setting the $\theta_i \in \{0, 0.05, 0.10, ..., 1.0\}$, providing a fine-grained analysis of the tracking performance across a wide range of overlap thresholds.

\subsection{Performance}
We tested different input modalities with SAM2 on the 2024 ICPR multi-modal tracking dataset. Including RGB image, infrared and depth. Among all the modalities, RGB achieve the highest performance. Results are shown in table \ref{table:multi-modal}. Incorporating SAM2 and our tricks resulted in 89.4 AUC score on the 2024 ICPR multi-modal tracking dataset, ranking 1st place among all the participants. 

\section{Conclusion}
In this paper, we presented an adaptation of the Segment Anything Model 2 for visual object tracking. By leveraging SAM2's advanced segmentation capabilities and integrating task-specific enhancements such as bounding box extraction, backward tracking, and tracklet interpolation. SAM2's inherent strength in maintaining object focus, even in challenging conditions like occlusion or motion blur, allowed us to achieve superior results on the 2024 ICPR Multi-modal Tracking challenge, securing first place with an AUC score of 89.4\%.

\section{Acknowledgement}
This work was supported by Electronics and Telecommunications Research Institute (ETRI) grant funded by the Korean government: 24ZD1120, Regional Industry ICT Convergence Technology Advancement and Support Project in Daegu-GyeongBuk (AI).

\bibliographystyle{splncs04}
\bibliography{ref}

\end{document}